\documentclass[10pt,twocolumn,letterpaper]{article}

\usepackage{cvpr}              

\usepackage[dvipsnames]{xcolor}

\DeclareUnicodeCharacter{2212}{-}

\usepackage{multirow}
\usepackage{subcaption}
\usepackage{multicol}

\definecolor{cvprblue}{rgb}{0.21,0.49,0.74}
\usepackage[pagebackref,breaklinks,colorlinks,citecolor=cvprblue]{hyperref}

\def\paperID{*****} 
\def\confName{CVPR}
\def\confYear{2024}

\title{An Empirical Study of Automated Mislabel\\Detection in Real World Vision Datasets}

\author{Maya Srikanth\thanks{Equal contribution.}, Jeremy Irvin\footnotemark[1], Brian Wesley Hill, Felipe Godoy, Ishan Sabane, Andrew Y. Ng\\
Stanford University\\
{\tt\small \{msrikant,jirvin16,bwhill,fgodoy,ishancs,ayn\}@stanford.edu}
}

\begin{document}

\maketitle

\newcommand{\MODEL}{SEMD}

\begin{abstract}

Major advancements in computer vision can primarily be attributed to the use of labeled datasets. However, acquiring labels for datasets often results in errors which can harm model performance. Recent works have proposed methods to automatically identify mislabeled images, but developing strategies to effectively implement them in real world datasets has been sparsely explored. Towards improved data-centric methods for cleaning real world vision datasets, we first conduct more than 200 experiments carefully benchmarking recently developed automated mislabel detection methods on multiple datasets under a variety of synthetic and real noise settings with varying noise levels. We compare these methods to a Simple and Efficient Mislabel Detector (\MODEL) that we craft, and find that \MODEL\ performs similarly to or outperforms prior mislabel detection approaches. 
We then apply \MODEL\ to multiple real world computer vision datasets and test how dataset size, mislabel removal strategy, and mislabel removal amount further affect model performance after retraining on the cleaned data. With careful design of the approach, we find that mislabel removal leads per-class performance improvements of up to 8\% of a retrained classifier in smaller data regimes.

\end{abstract}    
\section{Introduction}
Labeled datasets have driven immense progress in computer vision over the last decade \cite{imagenet,coco,laion, SAM}. Constructing such datasets often leads to labeling errors which can plague model behavior \cite{mislabel_effect_1, mislabel_effect_2, mislabel_effect_3}. This is especially relevant for application domains like healthcare and remote sensing, where highly specialized domain expertise can be required to accurately label images, but attaining these labels is costly compared to crowdsourcing or auto-labeling methods. Perhaps unsurprisingly, labeling errors have been shown to be highly prevalent in real world datasets \cite{northcutt2021pervasive,vasudevan2022does} and are known to be present even in datasets developed for crucial applications like medical diagnosis \cite{chexpert, mimic}. 

Towards addressing labeling errors in vision datasets automatically, many mislabel detection techniques have been developed in recent years \cite{cleanlab, simifeat, tracin, data_maps, AUM, characterizing, mentor_net, incv}. These methods show promise, but many have only been validated on datasets with synthetic noise and often tested with different evaluation protocols making them difficult to compare. Numerous works have explored developing mislabel detection and noisy label learning methods on real world vision datasets, but to the best of our knowledge, the applicability of recently developed mislabel detection methods to real world datasets has not been explored. These issues make it difficult for practitioners to understand how to implement recently developed mislabel detection approaches on their datasets. 

Furthermore, many of the methods examine training dynamics over an end-to-end, supervised training run in order to identify mislabeled examples for each task \cite{tracin, AUM,characterizing}, but advancements in pre-trained models present an opportunity to train a single model followed by much more efficient mislabel detection for many tasks. This has important implications in real world domains for which strong pre-trained models exist which are becoming increasingly prevalent\cite{bommasani2021opportunities,du2022survey}. Recent work has shown that running a k-nearest neighbor (kNN) algorithm on embeddings produced by a pre-trained encoder to identify mislabeled instances using their neighbor's distance-weighted labels leads to strong performance on multiple datasets with synthetic noise \cite{simifeat}, providing evidence of the potential of strong pre-trained models for improving mislabel detection.
However, it is still unclear how this and other mislabel detection methods can most effectively be applied to real world datasets.


To address these challenges, our primary contributions are as follows:
\begin{enumerate}
    \item To robustly evaluate newly developed mislabel detection methods, we conduct many experiments investigating their performance under a variety of synthetic and real label noise settings and across varying levels of label noise.
    \item We compare the methods to a simple and efficient mislabel detection approach that we design, which we call \MODEL\, that linear probes a pre-trained encoder and uses confident learning \cite{cleanlab} together with test-time augmentation on predictions generated by the resulting classifier to detect mislabels. The approach achieves state-of-the-art mislabel detection performance on CIFAR-10 and CIFAR-100 across most label noise settings and noise levels while being 5-20x faster than comparable approaches.
    \item We conduct a suite of additional synthetic label noise experiments using \MODEL\ to further quantify the importance of multiple design decisions including test-time augmentation and retraining procedure.
    \item Finally, we test the implementation of \MODEL\ in real-world, multi-label datasets for important applications (CheXpert \cite{chexpert}, METER-ML \cite{meter_ml}). 
    Specifically, we 
    systematically measure the impact of design decisions including multi-label removal strategy and amount of detected mislabels removed under various dataset sizes on performance of the classifier retrained on each cleaned dataset. We find that removing detected mislabels with \MODEL\ leads to substantial per-class improvements on smaller dataset sizes.

\end{enumerate}

To the best of our knowledge, no prior work has (a) carefully benchmarked recently developed mislabel detection approaches under a variety of synthetic noise settings and noise levels, (b) investigated the impact of multiple design decisions (performance tricks, retraining strategy) on mislabel detection development and application, and (c) systematically investigated the implementation of mislabel detection for real world datasets. We believe that our findings provide important insights into how to design effective and efficient mislabel detection approaches in practice. 

\section{Background}

\subsection{Datasets}

\textbf{CIFAR-10 and CIFAR-100} \, We use CIFAR-10 and CIFAR-100 \cite{cifar} as benchmarks for our synthetic noise experiments. Following prior work \cite{simifeat, tracin}, we introduce synthetic label noise into these datasets using three methods: symmetric noise (instance-independent), asymmetric noise (instance-independent), and confidence-based noise (instance-dependent), which are explained in Section~\ref{sec:synth_noise}. Additionally, we evaluate on the human-annotated versions CIFAR-10N and CIFAR-100N \cite{wei2022learning}, which simulate real-world noisy labels from data annotated by humans. 

\textbf{CheXpert and METER-ML} \, We use CheXpert \cite{chexpert} and METER-ML \cite{meter_ml} to investigate the effectiveness of our proposed approach on real-world datasets, where the type and prevalence of label noise is unknown.

CheXpert contains 224,316 chest X-rays labeled for the presence of 14 radiological observations. CheXpert contains mislabeled examples in its training set as it was labeled using an automatic labeler to extract mentions of observations from free-text radiology reports. However, the validation and test sets were labeled by a consensus of board-certified radiologists and therefore have high quality labels which can be used to robustly evaluate the performance of classifiers. 

METER-ML is a dataset of 86,599 multi-sensor images labeled for the presence of six methane-emitting facilities. The METER-ML training set consists of mislabeled data as the locations of infrastructure were obtained from public data sources which can be outdated and report shifted geographic coordinates. The validation and test sets, however, have high quality labels derived from a consensus of experts in methane emissions and related infrastructure. 

\subsection{Mislabel Detection Approaches}
We divide previous mislabel detection approaches into three categories: those based on using model predictions, model training dynamics, and model features.

\textbf{Prediction-Based:} \, Prior methods propose using predictions from a model trained on the noisy dataset to identify mislabeled data \cite{cleanlab, mentor_net, incv}. Notably, confident learning (CL) is a principled method with a highly used, open-source implementation (Cleanlab \cite{cleanlab_github}). CL uses model confidences to estimate a joint distribution of noisy (observed) labels and clean (latent) labels. The distribution is then used to estimate the number of mislabels in dataset partitions described by all pairs of noisy and clean labels, and prunes examples in these partitions according to rankings by model confidence.

\textbf{Training Dynamics:} \, Many recent mislabel detection approaches examine model training dynamics to deduce which examples are potentially mislabeled \cite{tracin, AUM,characterizing}. The Area Under the Margin (AUM) statistic measures the difference between the logits of the true class and the highest other class averaged over multiple epochs, and low values of AUM indicate likely mislabeled examples \cite{AUM}. Similarly, Second-Split Forgetting Time (SSFT) uses the epoch after which each example's label is never predicted correctly again (forgotten) after fine-tuning on a held-out subset of the data, and examples that are forgotten quickly are often mislabeled examples \cite{characterizing}. Another work developed a different approach called TracIn which traces the influence of each training example on its own loss during training using the magnitude of the gradient update. Examples with high self-influence are likely to be mislabeled \cite{tracin}. Importantly, each of these methods, although shown to accurately identify mislabeled examples, require an end-to-end training run for each task. 

\textbf{Feature-Based:} \,
A recently proposed method called SimiFeat clusters embeddings produced by a pre-trained encoder to identify likely mislabeled examples \cite{simifeat}. This is accomplished using a soft-label prediction via a kNN on the embeddings, where a sample's given label is compared to the soft-label generated from weighing the labels of nearby examples. As part of this process, SimiFeat uses a form of test-time augmentation (TTA), where the above kNN soft-classification process is repeated for many rounds with different random augmentations of each image, and an example is predicted to be mislabeled if it was voted to be mislabeled in over half the augmentation rounds. SimiFeat obtained state-of-the-art mislabel detection results on CIFAR-10 and CIFAR-100, but it is unclear whether the improved performance is primarily due to the method itself or the strength of the pre-trained encoder and the use of TTA. 

\begin{table*}[!ht]
    \centering
    \begin{tabular}{c|c|cccc|cccc}
    \hline
        \multirow{2}{*}{Approach} & Time &\multicolumn{4}{c|}{CIFAR-10} & \multicolumn{4}{c}{CIFAR-100} \\
        & GPU hr & Symm 0.6 & Asym 0.3 & Conf 0.1 & Human & Symm 0.6 & Asym 0.3 & Conf 0.1 & Human \\
        \hline
        Zero-shot \cite{clip} & 0.1 & 0.956 & 0.876 & 0.630 & 0.770 & 0.883 & 0.686 & 0.328 & \textbf{0.779} \\
        SimiFeat \cite{simifeat} & 2 & 0.968 & 0.896 & 0.801 & \textbf{0.884} & 0.881 & 0.723 & 0.498 & 0.770 \\
        TracIn* \cite{tracin} & 10 & \textbf{0.970} & 0.958 & \textbf{0.906} & \textbf{0.914} & \textbf{0.908} & 0.715 & \textbf{0.618} & \textbf{0.774}\\
        CL \cite{cleanlab} & 40 & 0.765 & \textbf{0.964} & 0.850 & 0.881 & 0.732 & \textbf{0.753} & 0.576 & 0.744 \\
        \hline
        \MODEL & 2 & \textbf{0.971} & \textbf{0.969} & \textbf{0.902} & 0.866 & \textbf{0.898} & \textbf{0.851} & \textbf{0.580} & 0.739 \\
        \hline
    \end{tabular}
    \caption{Mislabel detection results (F1) of \MODEL\ compared to baseline approaches on CIFAR-10 and CIFAR-100. All approaches use a CLIP-pre-trained ViT-B/32. We bold the top two performing methods in each column. GPU hours are measured on an NVIDIA RTX A4000 GPU.}
    \label{tab:baselines}
\end{table*}
\section{Experiments \& Results}

We describe the simple mislabel detection approach in Section~\ref{sec:model}, our synthetic mislabel experiments on CIFAR-10 and CIFAR-100 in Section~\ref{sec:synth_noise} and our real world mislabel experiments on CheXpert and METER-ML in Section~\ref{sec:real_world}.

\subsection{\MODEL: A Simple and Efficient Mislabel Detector}
\label{sec:model}

We aim to design a simple and efficient mislabel detection approach that can perform competitively with state-of-the-art approaches. We observe that prior state-of-the-art mislabel detection performance was achieved using a nearest neighbor approach on embeddings produced by a strong pre-trained image encoder (CLIP ViT-B/32 \cite{clip}) with TTA \cite{simifeat}. We hypothesize that this performance can be mostly attributed to the use of the strong encoder and TTA, and that task-specific training may further help identify mislabels (both of which our subsequent experiments support). For efficiency and to mitigate the impact of label noise, we opt to use linear probing (LP) of the frozen pre-trained encoder to obtain a classifier that is used for detecting the mislabeled examples with CL. To further improve efficiency and reduce dependency on GPUs, we implement the LP using logistic regression trained with LBFGS. Put together, the simple and efficient approach (1) uses logistic regression on embeddings from the CLIP image encoder to train a classifier, and (2) utilizes TTA along with the classifier to produce per-class probabilities and (3) applies CL to these probabilities to detect mislabeled examples. We refer to this \textbf{S}imple and \textbf{E}fficient \textbf{M}islabel \textbf{D}etection approach as \MODEL, visualized in Figure~\ref{fig:real-world-pipeline-multilabel} in the Supplementary Material.

\subsection{Synthetic Label Noise Experiments}
\label{sec:synth_noise}

We begin with experiments on the CIFAR-10 and CIFAR-100 datasets to compare the mislabel detection algorithms and assess the effect of multiple design decisions. Following prior work \cite{simifeat, tracin}, we introduce synthetic label noise into the datasets to measure mislabel detection performance and experiment with three types of noising procedures:

\begin{enumerate}[leftmargin=*]
    \item \textit{Symmetric Noise}: we change each sample's label to a different class selected uniformly at random.
    \item \textit{Asymmetric Noise}: we change all selected samples from each class to a fixed random class.
    \item\textit{Confidence-Based Noise}: we train a model on the clean dataset and use the highest scoring incorrect label as the noisy label.
    \item \textit{Human Noise}: we replace all labels with the labels given by a single human annotator \cite{wei2022learning}.
\end{enumerate}

In separate experiments, we apply each of the first three noising procedures to every image independently at a level $\eta \in [0, 1]$ ($\eta$ is the probability of changing its label under the procedure, $1$ - $\eta$ is the probability of retaining its given label), where we use $\eta=0.6$ for symmetric noise, $\eta=0.3$ for asymmetric noise, and $\eta=0.1$ for confidence-based noise following prior work \cite{simifeat}. We additionally experiment with varying this noise level under confidence-based noise as described in Section~\ref{sec:mislabel_detection}. The single annotator human noise has a known fixed error level of 17.23\% for CIFAR-10 and 40.20\% for CIFAR-100. The models are trained on the noisy labels then used to identify the potentially mislabeled images, and evaluated using the clean labels. We implement the logistic regression used in \MODEL\ with scikit-learn version 1.2.2 \cite{pedregosa2011scikit} using a weight decay of $0.001$. For training the deep learning models, we use an Adam optimizer with a weight decay of 1e−6, a batch size 32, and an initial learning rate of 0.001 for CIFAR-10 and 0.01 for CIFAR-100. We train for a maximum of 200 epochs and use early stopping with a patience of 10 epochs determined based on the accuracy on the validation set with clean labels. 

\subsubsection{Mislabel Detection}
\label{sec:mislabel_detection}

\textbf{Comparison of \MODEL\ with the Baseline Approaches}\, 
We compare \MODEL\ to several baseline mislabel detection approaches. We select the three representative and high-performing mislabel detection approaches, namely SimiFeat, TracIn, and CL using end-to-end fine-tuning. For confident learning, we use 10 rounds of cross-validation. For TracIn, we use 5 equally spaced model checkpoints and use gradients from the final layer to calculate self-influence, following the original work \cite{tracin}. As TracIn produces mislabel rankings but not binary mislabel indicators, we use thresholds produced by CL, following \cite{simifeat}. However, we note that TracIn is sensitive to choice of threshold and substantially benefits from the use of the CL threshold (see Figure~\ref{fig:tracin_cutoff} in the Supplementary Material), so we clearly indicate the use of this threshold in TracIn using `TracIn*'. For SimiFeat, we use a $k=10$ for the kNN and $M=21$ rounds of augmentation, following the original work \cite{simifeat}. For comparability, we use the CLIP-trained vision transformer (ViT-B/32) image encoder in all approaches. We additionally measure the total number of GPU hours taken by each approach, measured on a single NVIDIA RTX A4000 GPU.

As the CLIP vision-language model has demonstrated high performance on multiple benchmark datasets in a zero-shot regime, this naturally motivates the question of whether zero-shot classification with CLIP, by leveraging the jointly trained text encoder, would enable accurate mislabel identification performance without the influence of label noise. To assess this, we also include a zero-shot CLIP baseline where we replicate the zero-shot procedure proposed in \cite{clip} and compute the cosine similarity between the image embedding and text embedding ``a photo of a $<$\textit{class}$>$'' for each class. If the most similar class does not match its given label, we flag the image as mislabeled.

We find that \MODEL\ performs similarly to or outperforms all baseline approaches on six of the eight synthetic noise settings on CIFAR-10 and CIFAR-100 (Table ~\ref{tab:baselines}). TracIn* also obtains high performance on six of the eight noise settings, achieving performance within 0.01 F1 to \MODEL\ on most of the symmetric and confidence-based noise settings (excluding confidence-based noise on CIFAR-100 where it outperforms \MODEL\ by +0.038 F1). However, TracIn* underperforms \MODEL\ on the asymmetric noise settings (-0.011 and -0.136 F1 on CIFAR-10 and CIFAR-100 respectively), most prominently on CIFAR-100 where \MODEL\ substantially outperforms all other methods (+0.098 compared to TracIn). Notably, \MODEL\ is not in the top two performing methods under Human noise, which suggests that it makes predictions more similar to the human annotator than the other methods. CL performs similarly to or underperforms \MODEL\ across all synthetic noise settings on both datasets, mostly substantially underperforming on the symmetric noise settings (-0.206 and -0.156 F1 n CIFAR-10 and CIFAR-100 respectively), which may be explained by the end-to-end fine-tuning leading to overfitting of the noisy labels. \MODEL\ requires 5x and 20x less GPU time than the other consistently high performing approaches (TracIn* and CL respectively). 

The zero-shot CLIP approach underperforms most of the methods across all noise settings except for human noise on CIFAR-100, where it achieves the highest performance compared to all methods. It performs comparably to TracIn* and \MODEL\ on the symmetric noise settings, which could be due to its robustness to random label swapping, where images are often assigned labels from classes that are very different from their true class, but substantially underperforms the other methods on the asymmetric and confidence-based noise settings, which is likely due the fact that it embeds images of similarly appearing classes close to one another. We further find that the zero-shot approach performs very similarly to SimiFeat, with SimiFeat achieving +0.023 F1 higher score on average. This suggests that the CLIP encoder and TTA are mostly responsible for SimiFeat's strong performance, as opposed to the kNN procedure. However, this zero-shot approach does require a pre-trained text encoder, which is not always available depending on the image domain.

We quantify the impact of TTA on all methods and find that TTA improves the performance of all mislabel detection methods across all noise settings on both datasets (see Table~\ref{tab:tta} in the Supplementary Material) except for zero-shot with symmetric and human noise on CIFAR-100. The improvements are substantial for SimiFeat, leading to greater than 0.065 F1 improvements in all experiments. However, improvements are less pronounced for CL, TracIn*, and \MODEL\, with less than 0.03 F1 improvements across all experiments.

\begin{figure*}[htb]
    \centering

    \begin{subfigure}[t]{0.5\linewidth}
        \centering
        \label{fig:cifar10_mislabel_by_noise_rate}\includegraphics[height=3in]{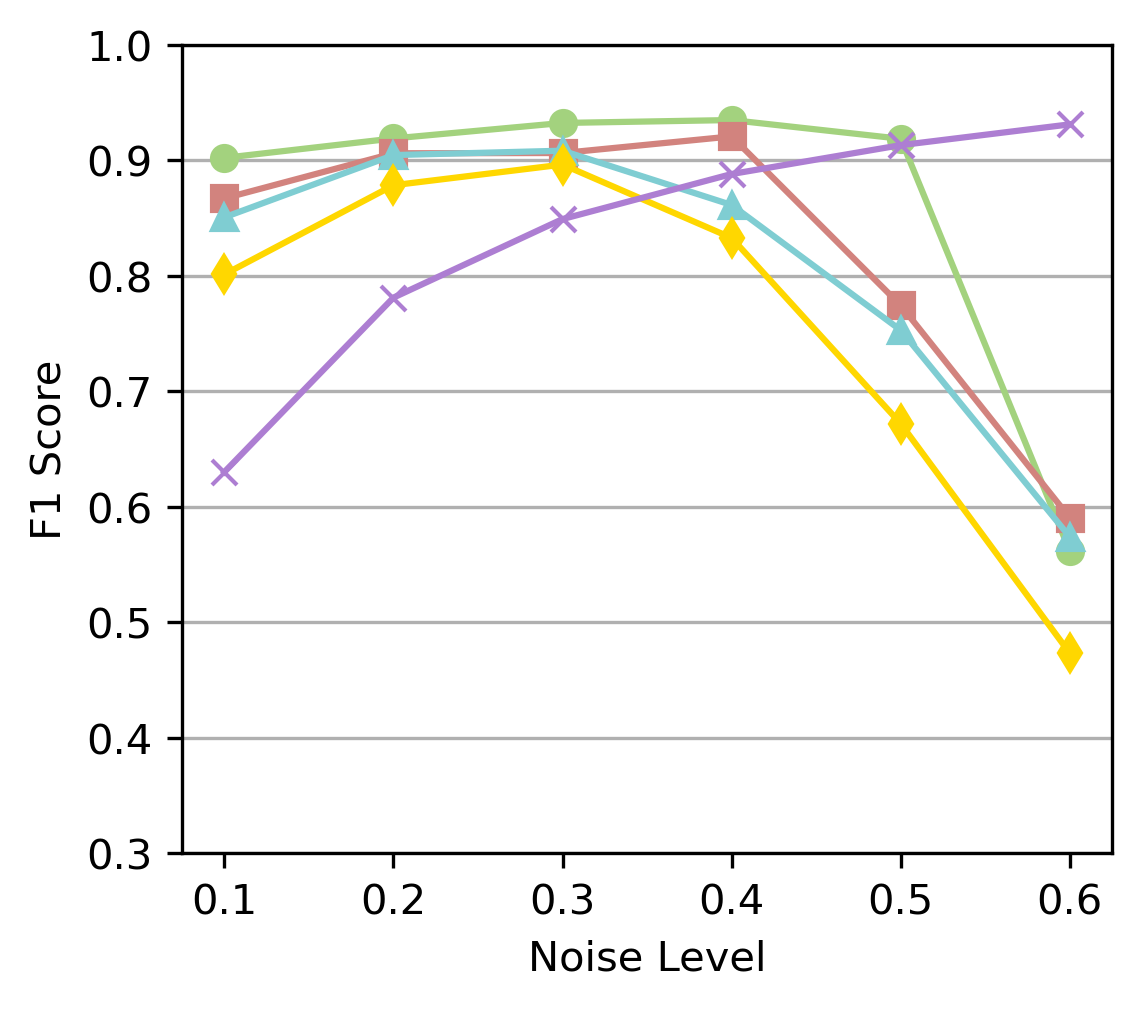}
        \caption{CIFAR-10}
    \end{subfigure}%
    ~
    \begin{subfigure}[t]{0.5\linewidth}
        \centering
        \label{fig:cifar100_mislabel_by_noise_rate}\includegraphics[height=3in]{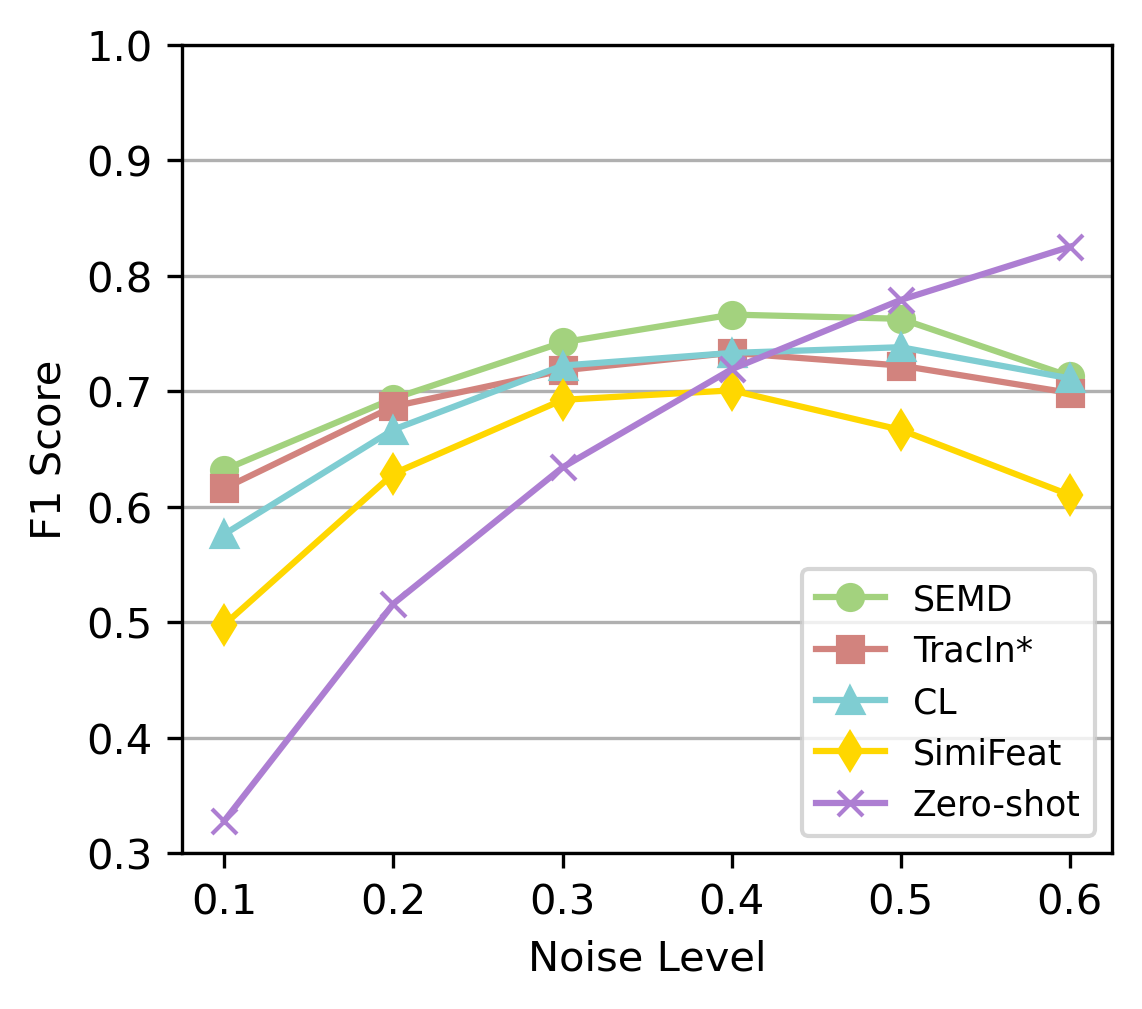}
        \caption{CIFAR-100}
    \end{subfigure}
    
    \caption{Mislabel detection results at varying levels of confidence-based noise on the training set of CIFAR-10 and CIFAR-100.}
    \label{fig:mislabel_by_noise_rate}
\end{figure*}

\textbf{Mislabel Detection Performance Across Varying Noise Levels} \,
It is possible that these trends in performance differences depend on the selected noise level, so we run experiments to compare the mislabel detection performance of \MODEL\ with the four baseline approaches across varying noise levels. Among the three noising procedures where we can control the noise level, we choose to run experiments with confidence-based noise as it is conditioned on the images (i.e. instance-dependent) and is therefore similar to human labeling mistakes, which are most likely to occur while differentiating two classes with similar appearances (e.g. crocodile vs. alligator). We vary the noise level in equal increments of 0.1 from 0.1 to 0.6 where the label noise affects the majority of examples in the dataset and substantially degrades performance.

We find that \MODEL\ achieves the highest mislabel detection performance across the 0.1 to 0.4 noise thresholds on both CIFAR-10 and CIFAR-100 (Figure~\ref{fig:mislabel_by_noise_rate}). It performs similarly to the zero-shot approach at a noise level of 0.5 but substantially underperforms the zero-shot approach at a noise level 0.6 where label noise becomes the majority. On CIFAR-10, TracIn* outperforms CL and SimiFeat across all noise levels but on CIFAR-100, it outperforms SimiFeat across all noise levels but performs similarly to CL (achieving slightly higher performance on 0.1 and 0.2 noise levels, very similar performance at 0.3 and 0.4, and slightly lower on 0.5 and 0.6). SimiFeat generally underperforms all methods on both datasets except for the zero-shot approach at 0.1-0.3 noise levels. The zero-shot approach performs best at high noise levels, substantially outperforming all methods at 0.6, which is likely due to the fact that the zero-shot is unaffected by label noise, unlike the other methods. The upward trend in zero-shot F1 score is due to increases in the prevalence of mislabels as the noise level is increased.

\begin{figure*}[t!]
    \centering
    \begin{subfigure}[t]{0.5\linewidth}
        \centering
        \includegraphics[height=3.0in]{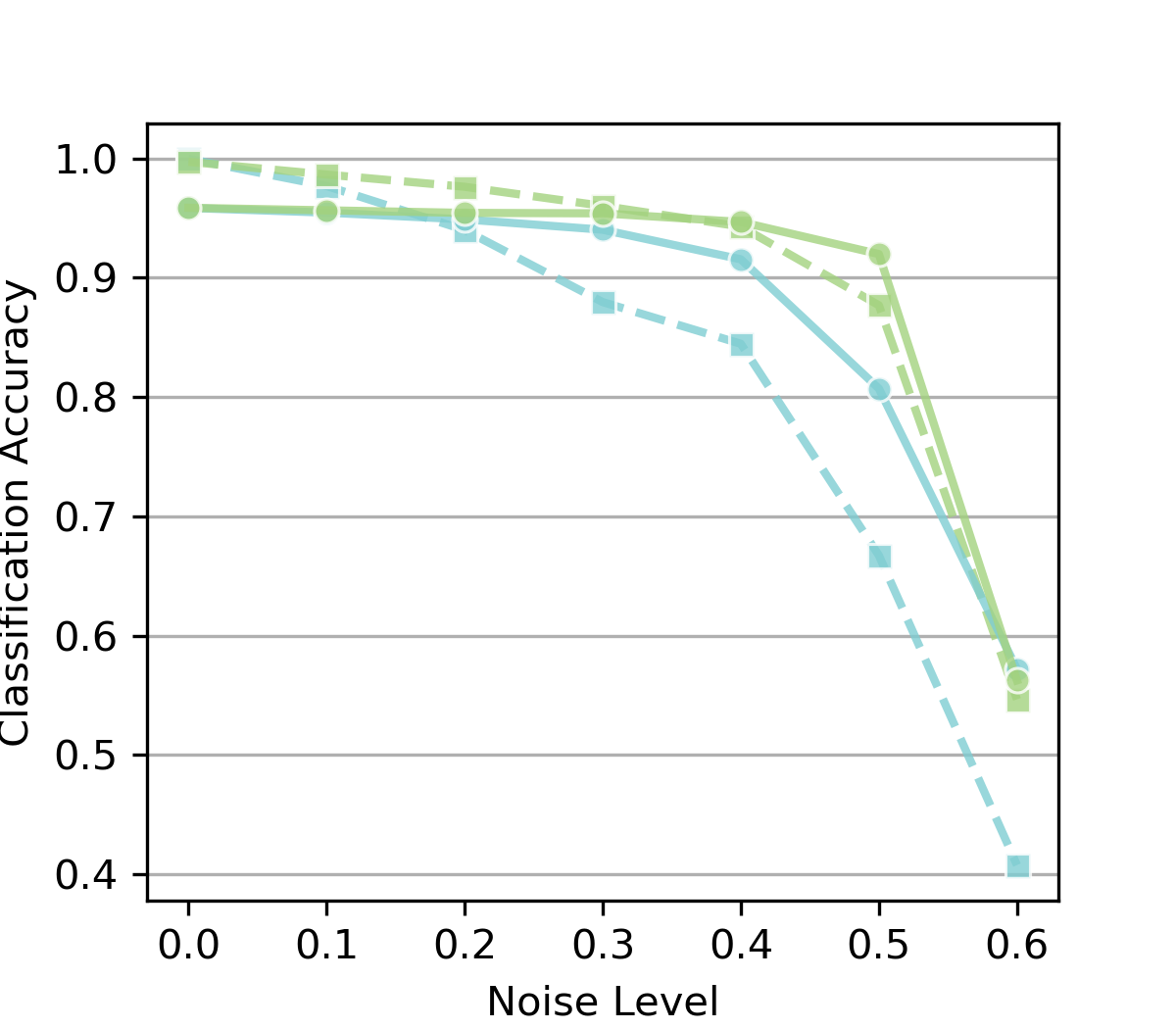}
        \label{fig:cifar10}
        \caption{CIFAR-10}
    \end{subfigure}%
    ~ 
    \begin{subfigure}[t]{0.5\linewidth}
        \centering
        \includegraphics[height=3.0in]{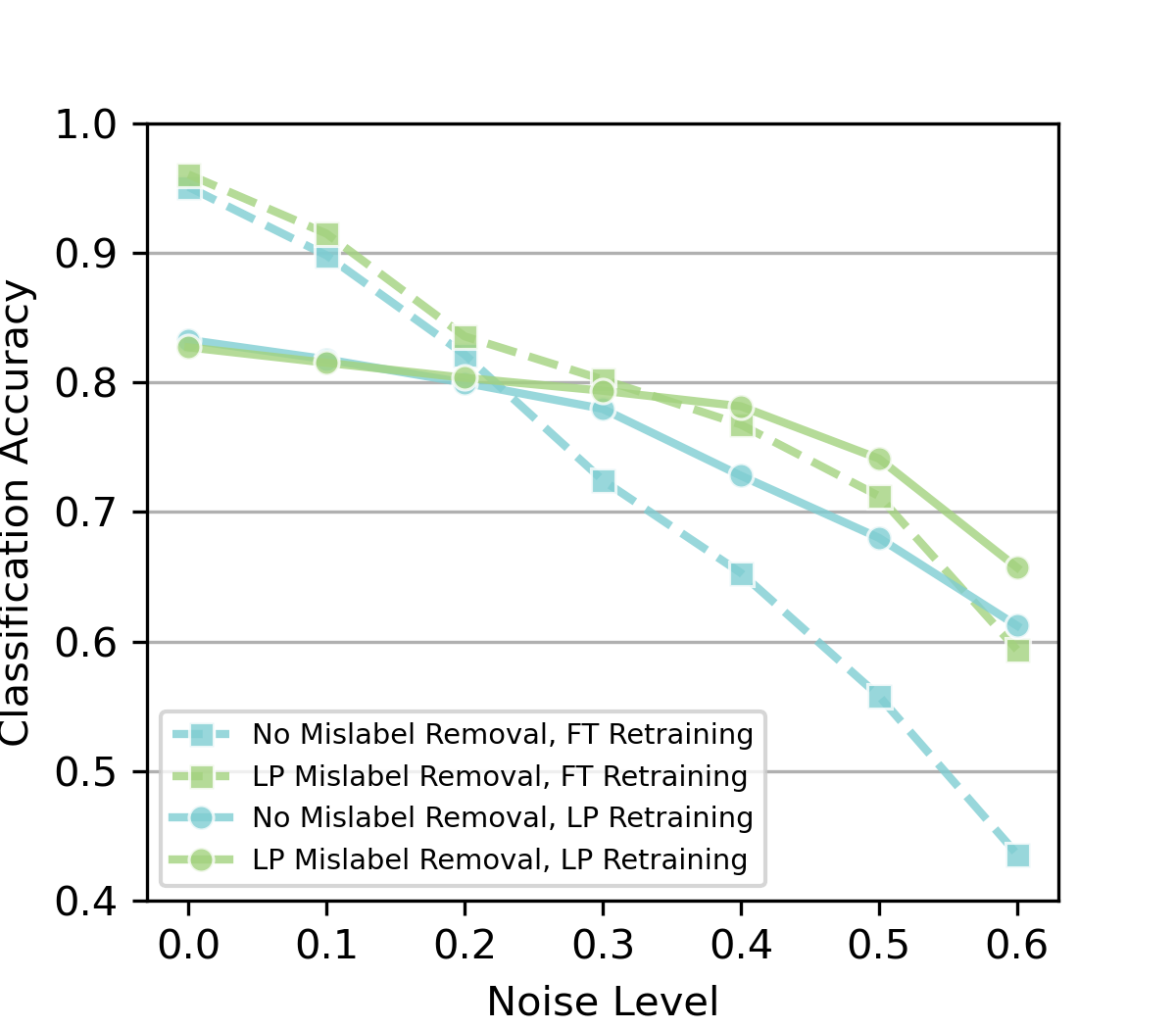}
        \label{fig:cifar100}
        \caption{CIFAR-100}
    \end{subfigure}

  \caption{Validation set performance of the retrained CLIP ViT-B/32 models using different fine-tuning strategies across varying levels of confidence-based noise on the training set of CIFAR-10 and CIFAR-100. We use \MODEL\ as the mislabel removal strategy for all experiments.}
  \label{fig:removal}
\end{figure*}

\subsubsection{Mislabel Removal and Retraining}
As the synthetic noise experiments suggest that \MODEL\ is an efficient and effective approach for mislabel detection, we proceed with that approach for all subsequent experiments.  To investigate the benefit of removing mislabels on the performance of a retrained classifier using \MODEL\, we run a variety of experiments evaluating the impact that various mislabel removal and retraining strategies have on the classification model’s performance. We explore this by first training the classification model on the original noisy training dataset, then running \MODEL\ on the training dataset to remove identified mislabeled examples, and finally training a new instance of the same model on the modified training dataset. As part of this exploration, we test the impact of the retraining procedure, specifically comparing linear probing to end-to-end fine-tuning when retraining with the identified mislabeled examples remove. We similarly conduct these experiments across varying label noise levels under confidence-based noise. We also measure the performance of linear probing and end-to-end fine-tuning on the original dataset without any examples removed as baselines.

Removing mislabels before retraining leads to performance improvements across almost all noise levels, with especially high improvements at intermediate noise levels (0.3 - 0.4) (Figure~\ref{fig:removal}). This may be explained by these noise levels being sufficiently large for mislabel removal to provide a meaningful difference in the classifier's performance while also being small enough to effectively develop the mislabel removal approach. At high noise levels (0.5 - 0.6), all approaches demonstrate a substantial drop in performance, but the mislabel removal still helps substantially except for linear probe retraining on CIFAR-10 at a noise level of 0.6.

LP retraining outperforms FT at higher noise levels on both datasets, whereas FT retraining outperforms LP at lower noise levels. FT retraining without mislabel removal demonstrates substantial ($>$ 10\%) drops in performance at intermediate to high noise levels on both datasets, which may be explained by its susceptibility to label noise memorization. LP retraining without mislabel removal is more robust to increases in label noise, but still underperforms mislabel removal before retraining on both datasets.

\subsection{Real World Mislabel Removal Experiments}
\label{sec:real_world}

Building upon the insights from our synthetic noise experiments, in this section we experiment with mislabel removal approaches on real world datasets. As the real world datasets are from a different domain than natural images, we select two high performing, domain-specific image encoders for each dataset to use as part of \MODEL. For CheXpert, we use the image encoder from GloRIA, a multi-modal contrastive image-text model trained on chest radiographic images and reports \cite{gloria}. For METER-ML, we use SatMAE, a masked autoencoder trained to reconstruct multi-spectral satellite imagery \cite{satmae}. On each dataset, we apply the \MODEL\ procedure where we use efficient linear probing (via logistic regression) on the encoder's output representations to obtain model probabilities, use these model probabilities for mislabel detection using CL, and then retrain again with linear probing on the training set with mislabels removed. We measure change in performance on a held-out test set (both overall and per-class) compared to a model trained on the training set with no mislabels removed. 

\subsubsection{Multi-Label Classification with Varying Amounts of Removed Mislabels}
The extension of this mislabel removal approach to real world datasets raises three key questions. First, many real world datasets are annotated as a multi-label task, where each example is labeled with a binary vector for the presence or absence of multiple independent classes, which requires changes to the mislabel detection approach. Second, many mislabel detection algorithms output a ranking of mislabeled examples, but it is not clear how to select the amount of mislabeled examples to remove from the dataset prior to retraining. This is further exacerbated in multi-class and multi-label settings, where due to differences in per-class noise levels, a different number of examples per class may need to be removed. For example, medical imaging datasets may contain disease classes that are more difficult to visually diagnose and are thereby more prone to label errors. Third, real world dataset sizes can vary substantially, and this may affect the impact of the mislabel detection approach on retraining performance. To shed light on these questions, we investigate how changing (1) the multi-label mislabel removal approach, (2) the amount of detected mislabels to remove, and (3) the dataset size each affect retraining performance.

\begin{figure*}[th!]
  \centering
  
  \begin{subfigure}{0.5\linewidth}
    \centering
    \includegraphics[height=2in]{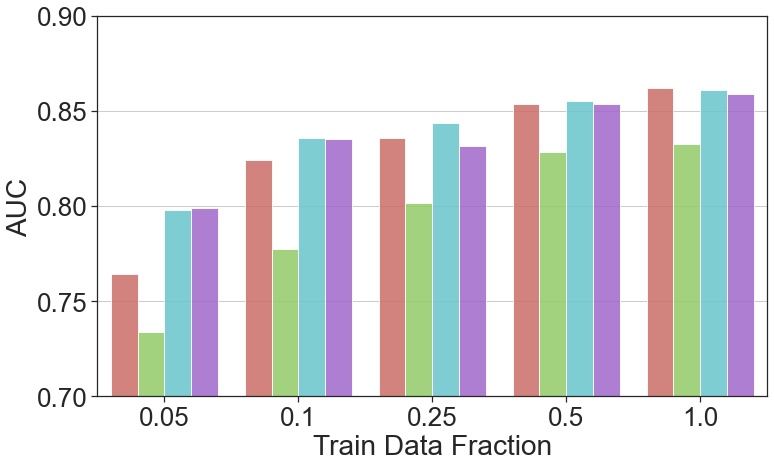}
    \label{fig:chexpert}
    \caption{CheXpert}
  \end{subfigure}%
  ~
  \begin{subfigure}{0.5\linewidth}
    \centering
    \includegraphics[height=2in]{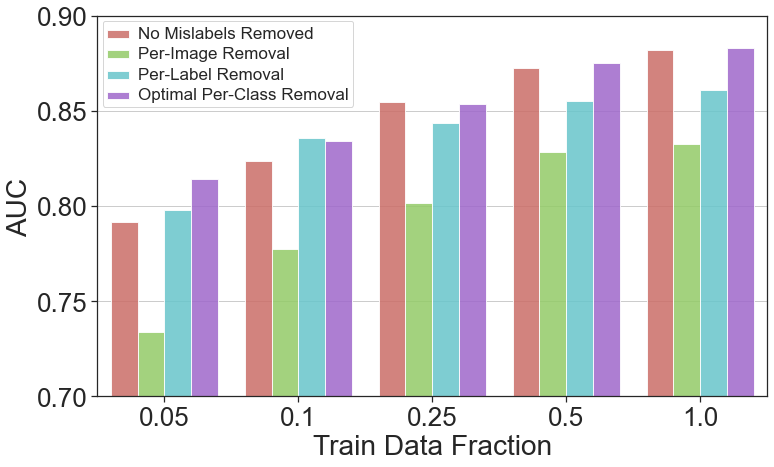}
    \label{fig:meter-ml}
    \caption{METER-ML}
  \end{subfigure}
  
    \caption{Average of the per-class AUC scores on the CheXpert (left) and METER-ML (right) test sets after retraining using the mislabel removal strategies with \MODEL\ on training datasets of varying sizes.}
    \label{fig:mml-chx-raw}
\end{figure*}

\textbf{Multi-Label Classification} \,
 Multi-label classification tasks present a unique challenge to automated mislabel detection: even with a small number of classes, there may be many possible subsets of clean labels and noisy labels for a single example. Accordingly, we consider two strategies for representing multi-label probabilities, each proposed in \cite{thyagarajan2022identifying}.
 The first strategy, which we refer to as `Per-Image Removal', estimates an aggregate label quality score for each multi-label example, flagging the example for removal if this score is low enough. This is the default approach implemented in Cleanlab \cite{cleanlab_github}. The second strategy, which we refer to as `Per-Label Removal', frames the multi-label classification task as multiple binary classification tasks, then uses CL to obtain task-specific mislabels. The first strategy discards entire examples with low enough removal scores, whereas the second strategy discards specific labels of examples but can retain examples even if they were identified to have mislabels.

 \textbf{Varying the Amount of Removed Mislabels} \,
 For each multi-label strategy, we experiment with varying the amount of mislabeled examples to remove. Specifically, we additionally experiment with a third strategy, which we refer to as `Optimal Per-Class Removal', which simply tries combinations of both per-image and per-label removal strategies with various amounts of mislabels removed \emph{for each class}, then selects the one that leads to the highest retraining performance on the validation set per-class. We note that this grid search is possible due to the computational efficiency of \MODEL. Finally, we compare Optimal Per-Class Removal with a model retrained on the original dataset with no mislabels removed. All methods are evaluated using the held-out test set of each dataset.

\textbf{Automated Mislabel Detection with Various Dataset Sizes} \,
Real world vision datasets can vary substantially in size which may affect the impact of mislabel detection approaches. To investigate this, we run the mislabel removal approaches using 5\%, 10\%, 25\%, 50\%, and 100\% of the CheXpert and METER-ML datasets respectively, and evaluate retraining performance on these fractions (see Table~\ref{tab:sample_sizes} in the Supplementary Material for the absolute number of examples corresponding to each fraction). 

\begin{figure*}[htb!]
  \centering
    \begin{subfigure}{1\linewidth}
      \includegraphics[ width=0.49\linewidth]{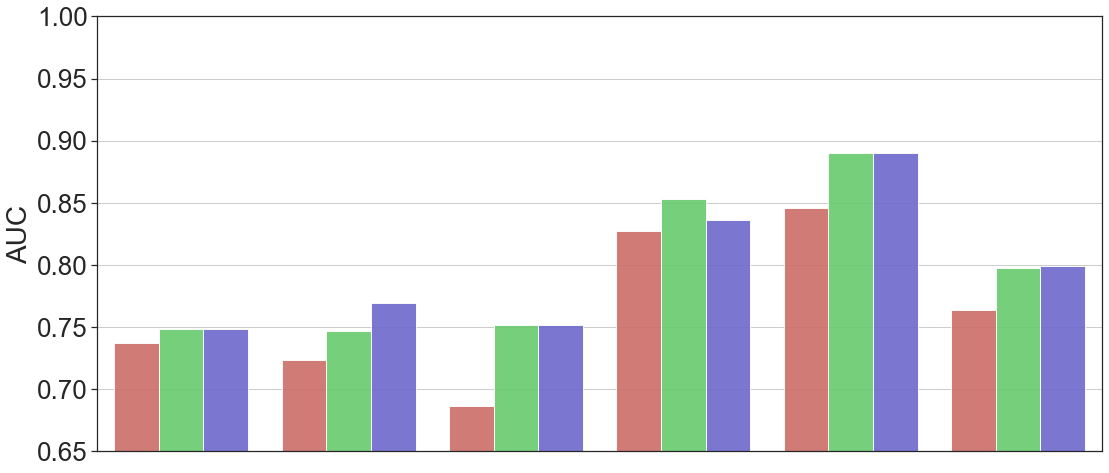}
      \includegraphics[width=0.49\linewidth]{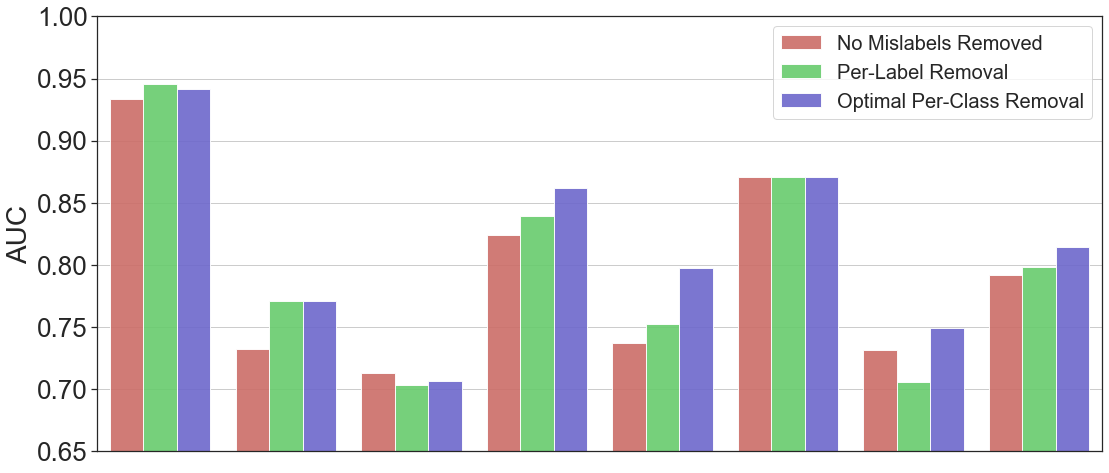}
      \vspace*{-1mm}
      \caption*{$5\%$}
    \end{subfigure}

    \begin{subfigure}{1\linewidth}
      \includegraphics[ width=0.49\linewidth]{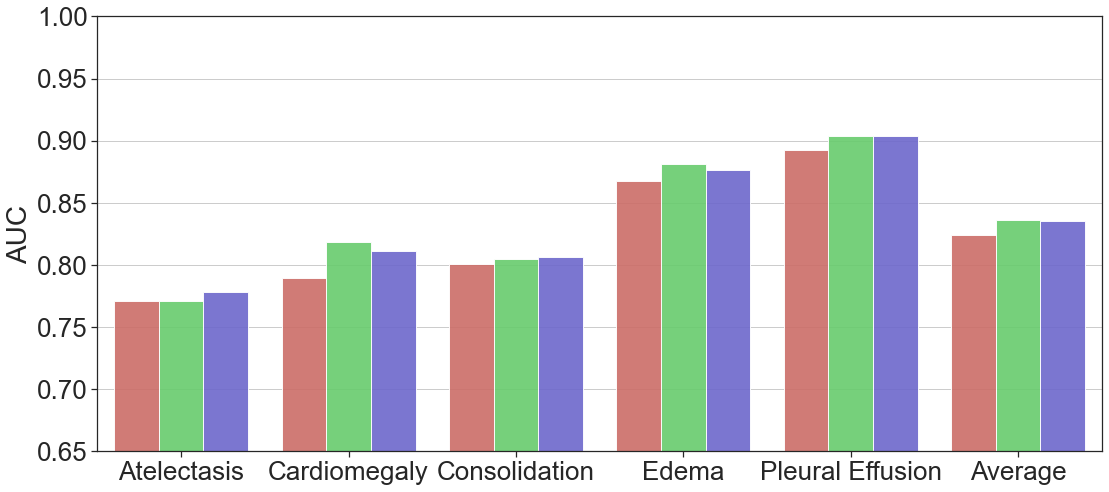}
      \includegraphics[width=0.49\linewidth]{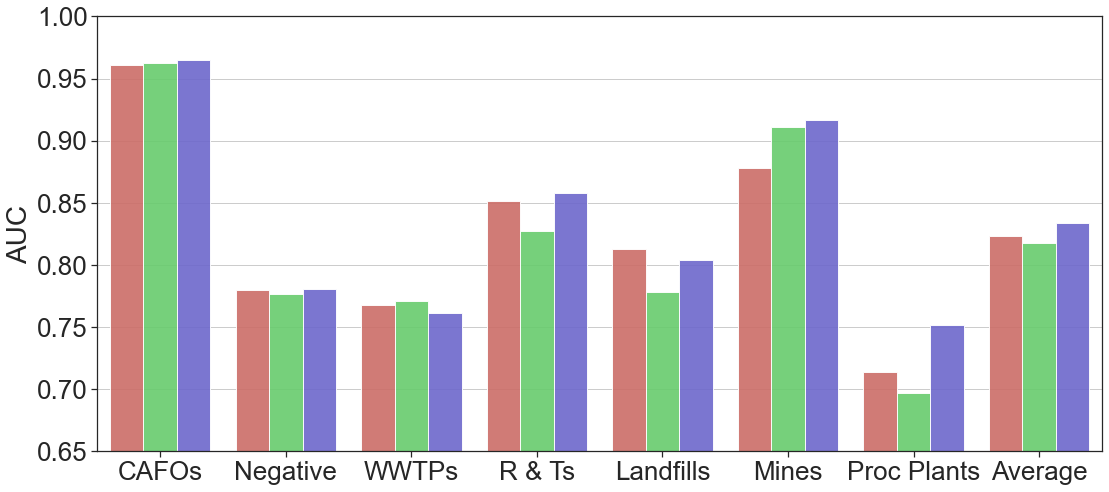}
      \vspace*{-1mm}
      \caption*{$10\%$}
    \end{subfigure}

  \caption{Per-class AUC differences on CheXpert (left) and METER-ML (right) across lower data regimes after using \MODEL. The per-class performance at higher data regimes is in Figure~\ref{fig:mml-chx-per-class-BIG} in the Supplementary Material.}
  \label{fig:mml-chx-per-class}
\end{figure*}
 
\textbf{Results} \, We find that the Optimal Per-Class Removal strategy leads to a retrained model which either performs better than (or on par with) Per-Label Removal on both datasets (Figure~\ref{fig:mml-chx-raw}). The Per-Image Removal removal strategy performs poorly out of the box, both overall and at the per-class level. We suspect this is because the strategy is overconfident in \emph{automated} mislabel removal, as it discards entire examples based on an aggregate label quality score rather than discarding specific labels. Per-Label Removal performs comparably with Optimal Per-Class Removal, except in the lower data regimes of METER-ML, where Optimal Per-Class Removal outperforms the other two strategies. Furthermore, the Optimal Per-Class Removal procedure gives the highest returns in low data regimes of $\sim 4k$ to $\sim 20k$ examples, with diminishing returns as the size of the data increases. We hypothesize this trend is likely because (1) models trained on larger datasets are less sensitive to noisy examples, as has been demonstrated in prior work \cite{zhang2021understanding, sajjadi2016regularization, lee2018cleannet, han2019deep} and (2) the mislabel detection methods are still imperfect, and removed examples may actually be correctly labeled or otherwise informative to the training process.




The per-class performance differences follow similar trends to the aggregate performance, where mislabel removal at lower data fractions leads to more substantial improvements per-class (Figure \ref{fig:mml-chx-per-class}). For example, in the 5\% data regime of CheXpert, \MODEL\ improves performance on Cardiomegaly and Atelectasis by 3.27 and 3.04 AUC respectively. However, it only improves performance on Consolidation by 0.26 AUC and even slightly decreases the performance of Edema by 0.46 AUC. Furthermore, in the 5\% data regime of METER-ML, \MODEL\ leads to improvements above no mislabel removal across almost all classes, notably including an improvement of 6.05 AUC on Landfills. However, on waste water treatment plants (WWTPs) it leads to a small decrease (-0.01 AUC) in performance. The per-class differences at higher data regimes are minimal on both datasets, where sometimes mislabel removal leads to small improvements and sometimes it leads to small reductions in performance (see Figure~\ref{fig:mml-chx-per-class-BIG} in the Supplementary Material). Notably, on CheXpert, performance on Consolidation seems to be harmed by mislabel removal across the 25\%, 50\%, and 100\% regimes. Generally the Optimal Per-Class Removal performs on par with Per-Label Removal across the higher data regimes, but the former strategy generally outperforms the latter on the lower data regimes.
\section{Discussion}
Through more than 200 experiments, we find that a simple and efficient approach outperforms a variety of strong baselines across most tested synthetic noise settings and noise levels on CIFAR-10 and CIFAR-100. We further find that design decisions including TTA, retraining procedure, amount of removed mislabels, and mislabel removal strategies are important to the resulting model performance. Finally, we apply the simple and efficient mislabel detector to two real world datasets and find benefits under lower data regimes with particularly high improvements on specific classes. Our work provides insights into how to develop an effective and efficient mislabel detection approach for real world datasets.



Our work has limitations. First, although we demonstrate the feasibility of designing effective automated mislabel detection methods which improve the performance of model upon retraining, their performances do not reach the performance of a model trained on perfectly labeled data. Practitioners should still try to ensure quality data collection and labeling.
Second, although linear probing worked well in our experiments, it may underperform in settings with large distribution shifts \cite{lpft,lee2022surgical}.
Third, we neither aim to specifically identify the impact of nor explicitly limit noise label memorization in this work, which prior works have shown can be mitigated with early stopping \cite{rolnick2017deep,li2020gradient,xia2021robust} and regularization \cite{liu2020peer,liu2020early,cheng2020learning,cheng2021mitigating}.

We believe there are several interesting directions for future work. First, we only explore automatically removing mislabels, but future work should consider automatically relabeling mislabeled data which may be advantageous. Second, we only focus on classification tasks in this work, but the CL methods can be extended to any task that Cleanlab supports, including object detection and semantic segmentation \cite{cleanlab}. Extending to vision-language tasks, however, is more challenging. Third, although we experiment with varying noise rates under confidence-based noise, we do not do so for symmetric and asymmetric noise. It may be worthwhile for future work to explore this.

{
    \small
    \bibliographystyle{ieeenat_fullname}
    \bibliography{bibliography}
}

\clearpage
\setcounter{page}{1}
\maketitlesupplementary

\section{Supplementary Material}


\subsection{Sensitivity of TracIn F1 Score to Confident Learning Threshold}
\label{sec:tracin_sensitivity}
\begin{figure*}[htb!]
    \centering
    
    \begin{subfigure}{0.3\linewidth}
        \label{fig:symm}\includegraphics[height=1.2in]{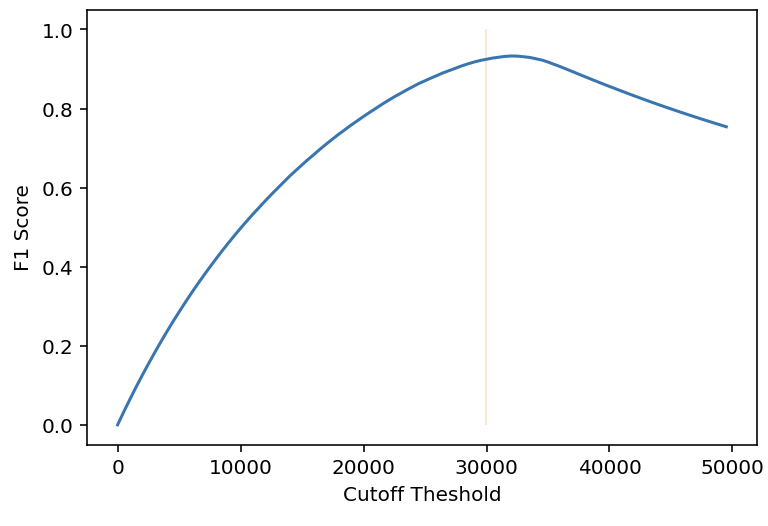}
        \caption{Symm 0.6}
    \end{subfigure}
    ~
    \begin{subfigure}{0.3\linewidth}
        \label{fig:asymm}\includegraphics[height=1.2in]{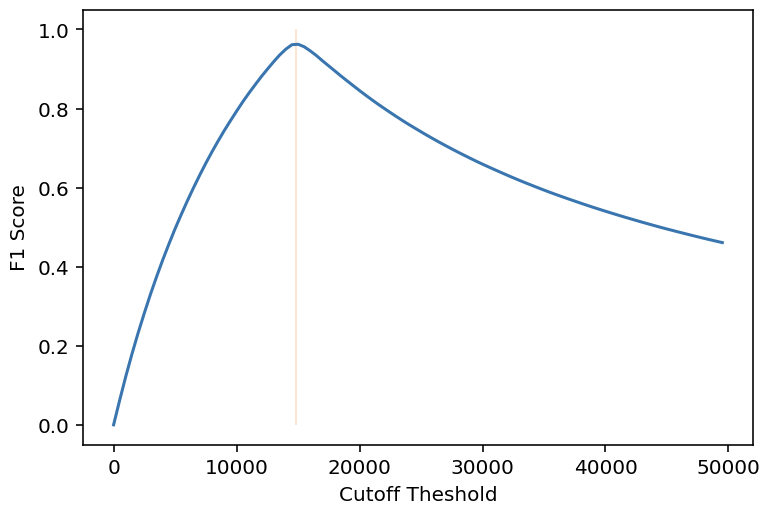}
        \caption{Asym 0.3}
    \end{subfigure}
    ~
    \begin{subfigure}{0.3\linewidth}
        \label{fig:adv}\includegraphics[height=1.2in]{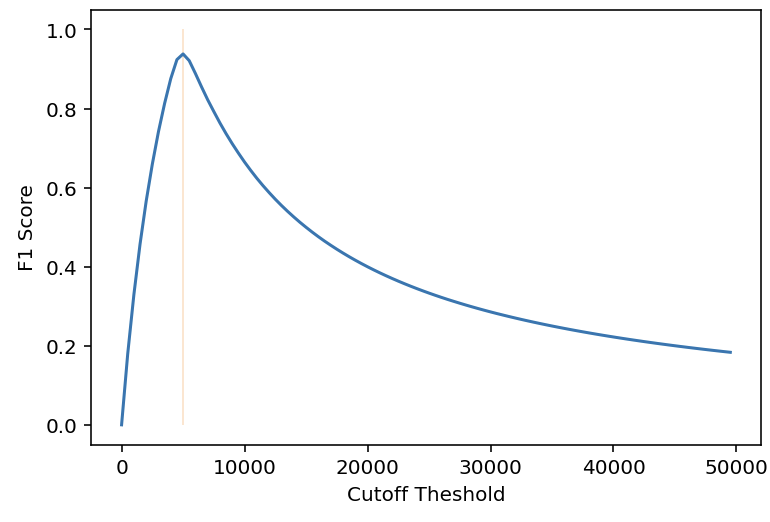}
        \caption{Conf 0.1}
    \end{subfigure}

    \caption{Importance of the cutoff threshold on mislabel detection F1 score with TracIn. Orange lines represent the end-to-end finetuning confident learning threshold, which is used in our F1 score calculations.}
    \label{fig:tracin_cutoff}
\end{figure*}

TracIn outputs a ranking over all the examples in the dataset, so setting a threshold is required to evaluate the performance of the mislabel detector in a binary evaluation setting. Following \cite{simifeat}, we use the CL end-to-end fine-tuned model's computed number of identified mislabels to threshold TracIn, and refer to this approach as TracIn*. To investigate the importance of this threshold selection, we compute the mislabel F1 score at various mislabel number thresholds (Figure~\ref{fig:tracin_cutoff}). We find that performance is highly sensitive to the choice of threshold, with F1 score dropping substantially if the threshold is set too high or too low. Importantly, the confident learning thresholds are very close to the F1 score maximizing threshold, indicating that TracIn* F1 scores may be buoyed by additional information from CL. 

\subsection{Summary of \MODEL}

\begin{figure*}[htb!]
    \centering
    \includegraphics[width=0.85\textwidth]{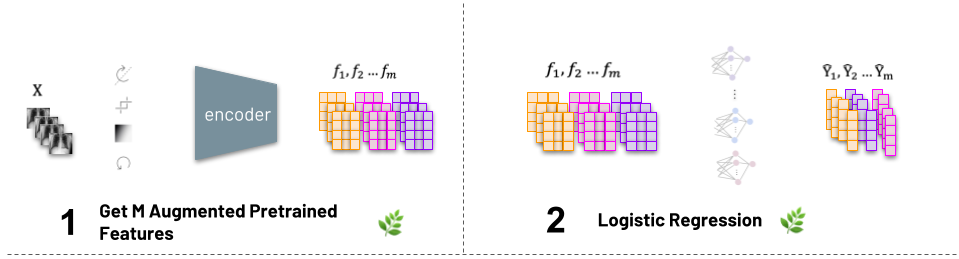}  \\
    \includegraphics[width=0.85\textwidth]{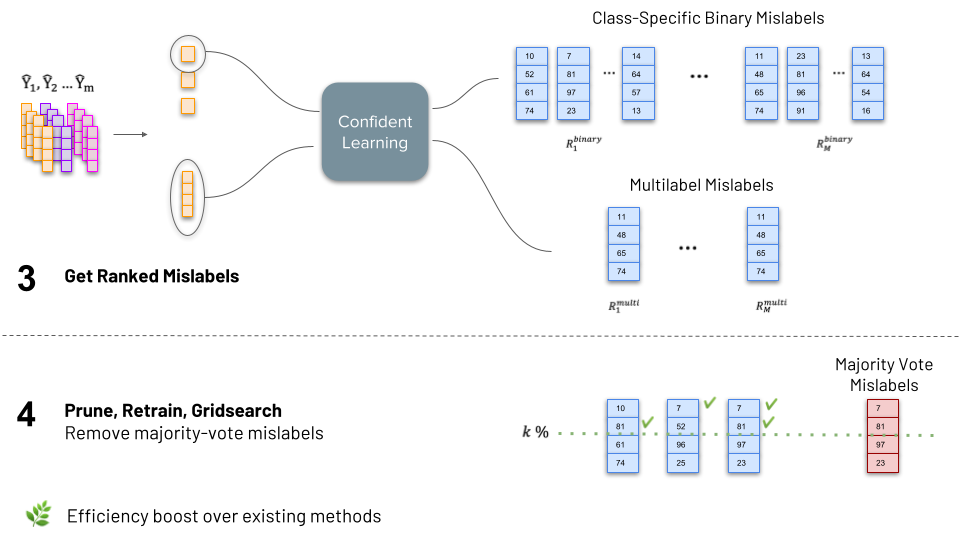}
    \caption{Efficient, Real-world Mislabel Removal for Multi-Label Data. We synthesize findings from earlier ablations to offer an efficient, 4-step pipeline for mislabel detection on multi-label data which leverages pre-trained features, fast logistic regression, test-time augmentation, and gridsearching over important mislabel detection hyperparameters. Note that gridsearching process (step 4) assumes access to clean validation and test sets.}
    \label{fig:real-world-pipeline-multilabel}
\end{figure*}

\MODEL\ can be applied to datasets of any size by obtaining pre-trained features from an in-domain encoder, training a logistic regression model to predict the labels given these pre-trained features, obtaining and removing the mislabels using confident learning with the trained logistic regression model, and finally retraining on the dataset with mislabels removed. Figure \ref{fig:real-world-pipeline-multilabel} shows this mislabel removal pipeline for real-world data: note that the speed of logistic regression on pre-trained features enables gridsearching over many settings of the relevant mislabel detection hyperparameters. 

\begin{table*}[!ht]
    \centering
    \begin{tabular}{cc|cccc|cccc}
    \hline
        \multirow{2}{*}{Approach} & \multirow{2}{*}{TTA} &\multicolumn{4}{c|}{CIFAR-10} & \multicolumn{4}{c}{CIFAR-100} \\
        & & Symm 0.6 & Asym 0.3 & Conf 0.1 & Human & Symm 0.6 & Asym 0.3 & Conf 0.1 & Human \\
        \hline
        \multirow{2}{*}{Zero-shot} & - & 0.956 & 0.876 & 0.630 & 0.770 & 0.883 & 0.686 & 0.328 & 0.779 \\
        & \checkmark & 0.962 & 0.891 & 0.758 & 0.767 & 0.869 & 0.714 & 0.433 & 0.578\\
        \hline
        \multirow{2}{*}{CL} & - & 0.765 &	0.964 & 0.850 & 0.881 & 0.732 & 0.753 & 0.576 & 0.744\\
        & \checkmark & 0.779 & 0.973 & 0.881 & 0.894 & 0.748 & 0.769 & 0.604 & 0.751 \\
        \hline
        \multirow{2}{*}{TracIn*} & - & 0.969 & 0.957 & 0.907 & 0.914 & 0.908 & 0.716 & 0.618 & 0.774\\
        & \checkmark & 0.974 & 0.964 & 0.920 & 0.917 & 0.915 & 0.725 & 0.632 & 0.777 \\
        \hline
        \multirow{2}{*}{SimiFeat} & - & 0.870 & 0.755 & 0.731 & 0.828 & 0.798 & 0.614 & 0.433 & 0.711\\
        & \checkmark & 0.968 & 0.896 & 0.801 & 0.884 & 0.881 & 0.723 & 0.498 & 0.770\\
        \hline
        \multirow{2}{*}{\MODEL} & - & 0.961 & 0.957 & 0.872 & 0.849 & 0.881 & 0.824 & 0.544 & 0.727\\
        & \checkmark & 0.971 & 0.969 & 0.902 & 0.866 & 0.898 & 0.851 & 0.580 & 0.739\\
        \hline
    \end{tabular}    
    \caption{Effect of test-time augmentation (TTA) on the mislabel detection performance of all approahces. We use 21 augmented images when using TTA. All approaches use CLIP ViT-B/32 as the encoder.}
    \label{tab:tta}
\end{table*} 

\begin{table*}[htb!]
    \centering
    \begin{tabular}{ccc}
    \hline
    Fraction & CheXpert & METER-ML \\
    \hline
    $5\%$  & 4,300 & 9,600 \\
    $10\%$ & 8,600 & 19,000 \\
    $25\%$ & 21,500 & 48,000 \\
    $50\%$ & 44,000 & 96,000 \\
    $100\%$ & 87,000 & 192,000 \\
    \hline
    \end{tabular}
    \caption{Approximate number of examples included for real world datasets at the various fractions of data.}
    \label{tab:sample_sizes}
\end{table*}

\begin{figure*}[htb!]
  \centering
    \begin{subfigure}{1\linewidth}
      \includegraphics[ width=0.49\linewidth]{figures/chexpert_results/chexpert_perclass_test_AUC_frac_0.05.png}
      \includegraphics[width=0.49\linewidth]{figures/meterml_results/meter_ml_perclass_test_AUC_frac_0.05.png}
      \vspace*{-1mm}
      \caption*{$5\%$}
    \end{subfigure}

    \begin{subfigure}{1\linewidth}
      \includegraphics[ width=0.49\linewidth]{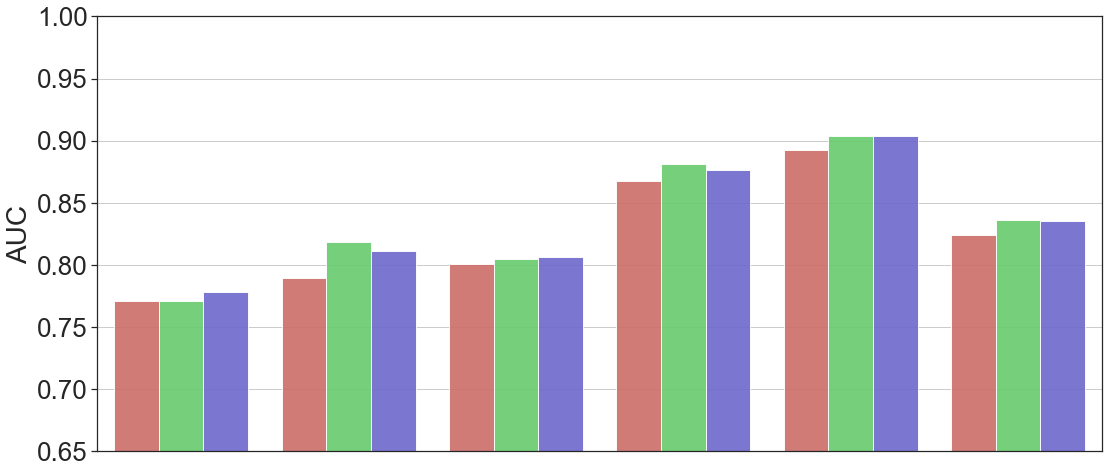}
      \includegraphics[width=0.49\linewidth]{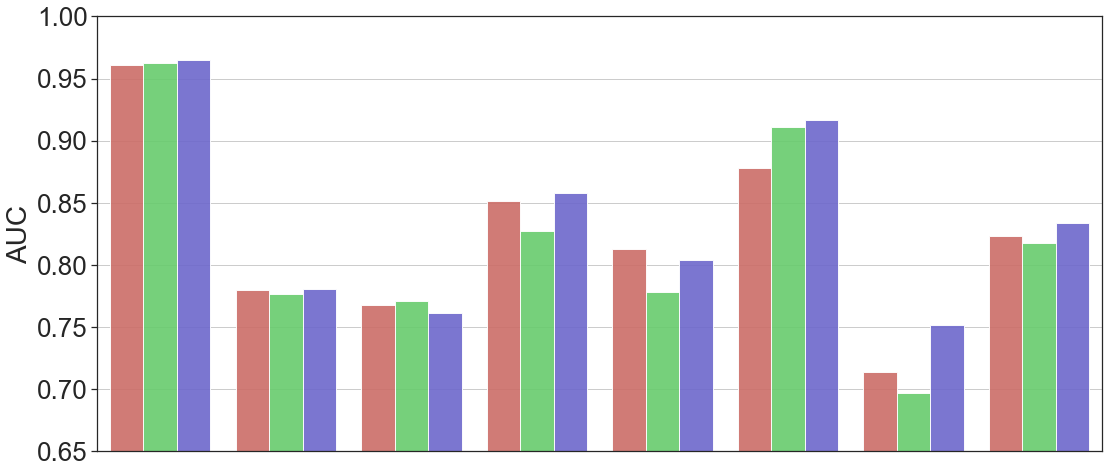}
      \vspace*{-1mm}
      \caption*{$10\%$}
    \end{subfigure}

    \begin{subfigure}{1\linewidth}
      \includegraphics[ width=0.49\linewidth]{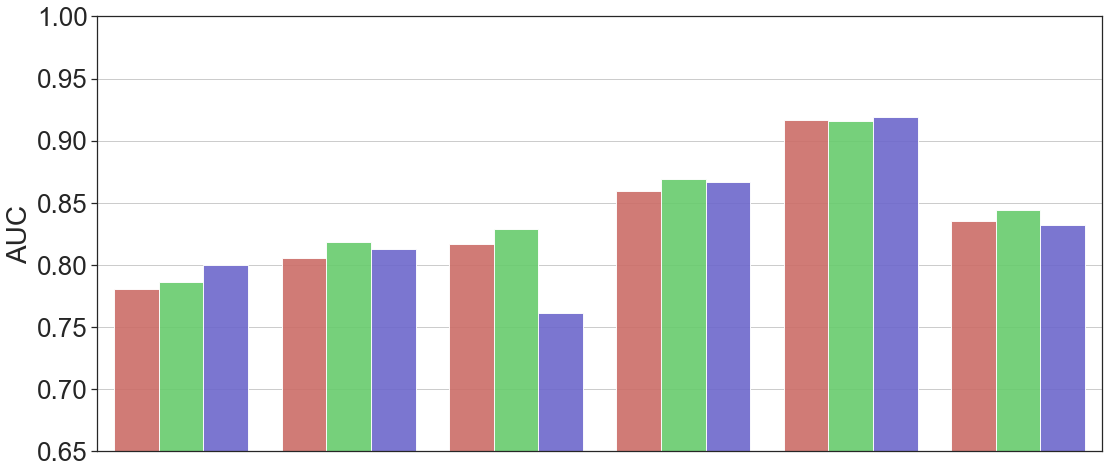}
      \includegraphics[ width=0.49\linewidth]{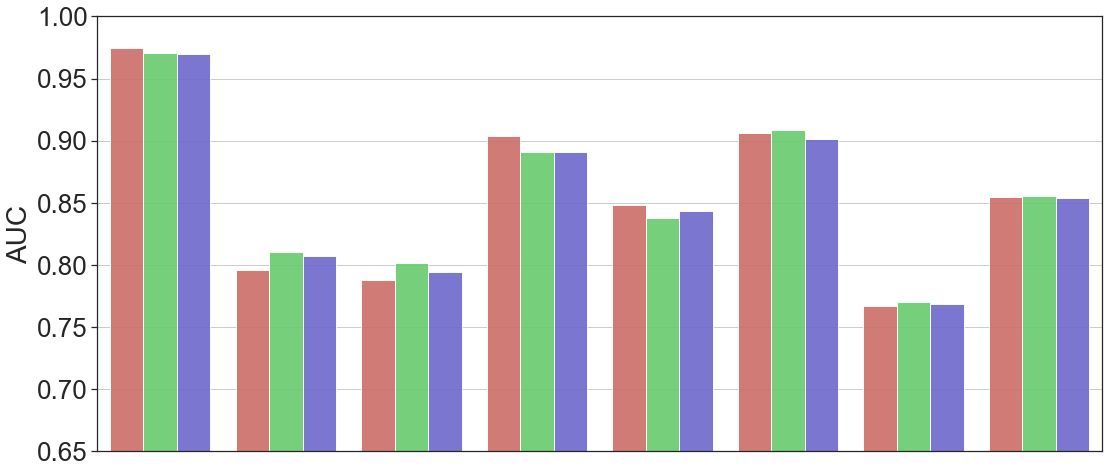}
      \vspace*{-1mm}
      \caption*{$25\%$}
    \end{subfigure}    

    \begin{subfigure}{1\linewidth}
      \includegraphics[ width=0.49\linewidth]{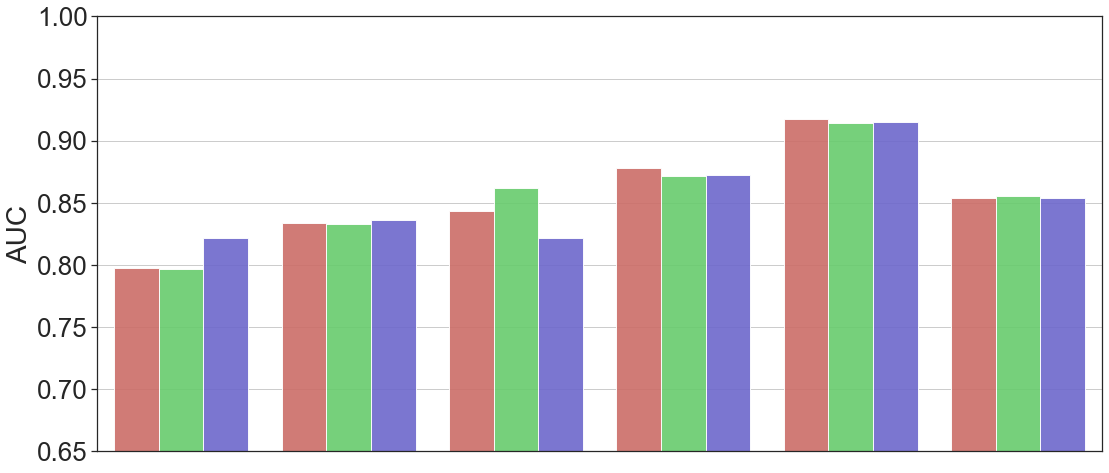}
      \includegraphics[ width=0.49\linewidth]{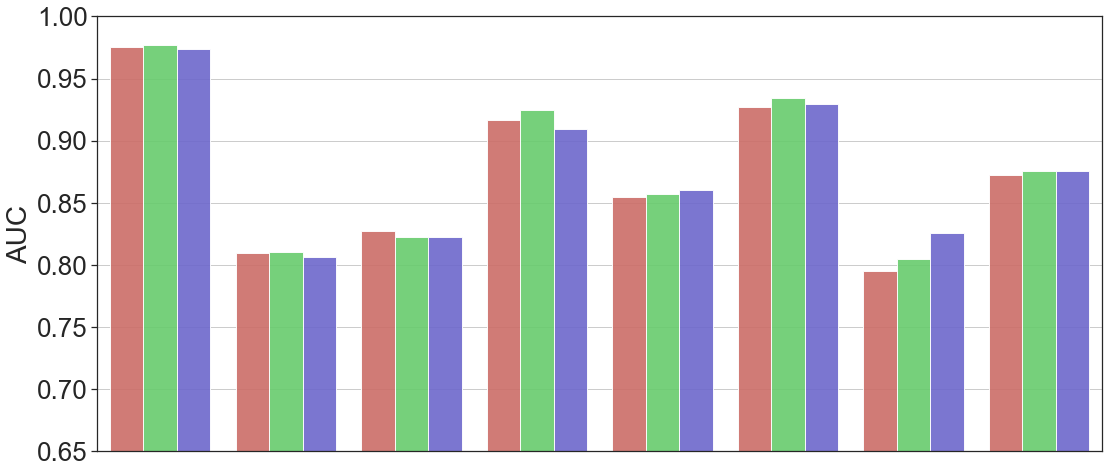}
      \vspace*{-1mm}
      \caption*{$50\%$}
    \end{subfigure}    

    \begin{subfigure}{1\linewidth}
      \includegraphics[ width=0.49\linewidth]{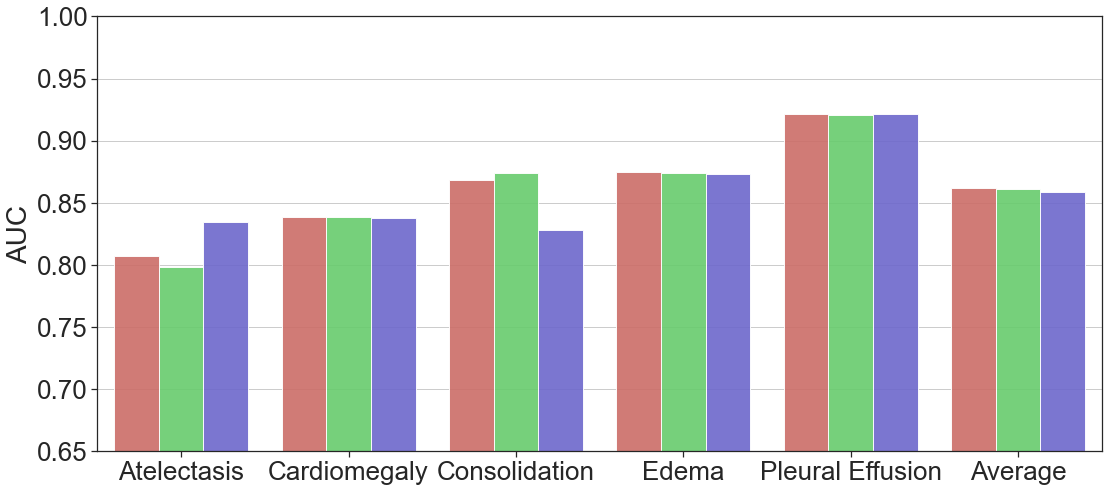}
      \includegraphics[ width=0.49\linewidth]{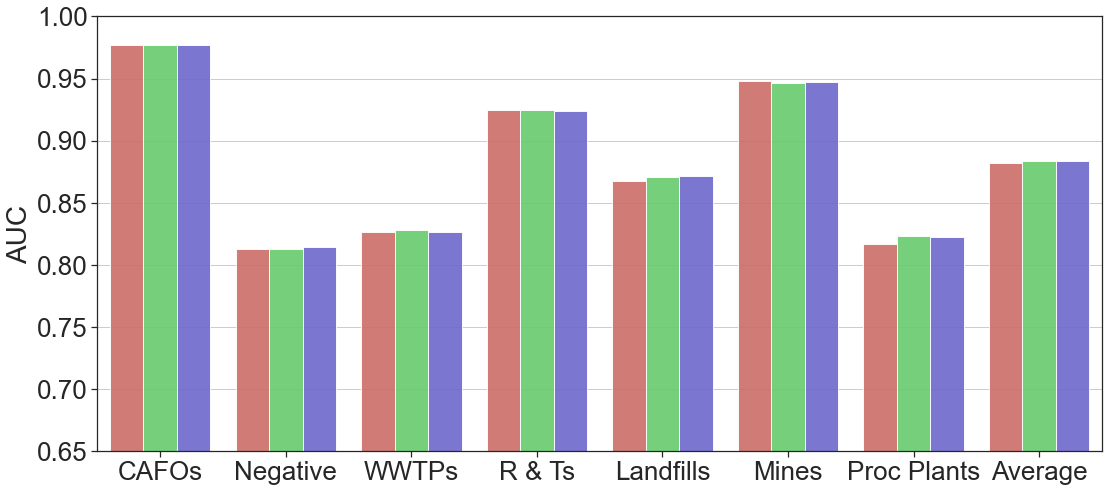}
      \vspace*{-1mm}
      \caption*{$100\%$}
    \end{subfigure}  

  \caption{Per-class AUC change on CheXpert (left) and METER-ML (right) cross various data regimes after using \MODEL. Superset of the results in Figure~\ref{fig:mml-chx-per-class}.}
  \label{fig:mml-chx-per-class-BIG}
\end{figure*}

\end{document}